%% file: main.tex
\documentclass[10pt,twocolumn,letterpaper]{article}

\usepackage[pagenumbers]{cvpr}

\input{preamble}
\definecolor{cvprblue}{rgb}{0.21,0.49,0.74}
\usepackage[pagebackref,breaklinks,colorlinks,allcolors=cvprblue]{hyperref}

\def\paperID{33623}
\def\confName{CVPR}
\def\confYear{2026}

\title{Learning from Noisy Prompts:

Saliency-Guided Prompt Distillation for Robust Segmentation with SAM}

\author{
Jingxuan Kang$^{1}$\quad
Ziqi Zhang$^{1}$\quad
Shaoming Zheng$^{1}$\quad
Shuang Li$^{2}$\\
Uday Bharat Patel$^{3}$\quad
Alexander Harry Fitzhugh$^{3}$\quad
Phillip Lung$^{3}$\quad
Yusuf Kiberu$^{3}$\\
Nikesh Jathanna$^{4}$\quad
Shahnaz Jamil-Copley$^{3,4}$\quad
Bernhard Kainz$^{1}$\quad
Chen Qin$^{1}$\thanks{Corresponding author.}\\[2pt]
{\small $^{1}$Imperial College London\quad
$^{2}$Beihang University\quad
$^{3}$National Health Service\quad
$^{4}$University of Nottingham}
}

\begin{document}
\maketitle
\input{Chapters/0_Abstract}
\input{Chapters/1_Introduction}
\input{Chapters/2_Related_Work}
\input{Chapters/3_Method}

\input{Chapters/4_Experiment}
\input{Chapters/5_Conclusion}

{
    \small
    \bibliographystyle{ieeenat_fullname}
    \bibliography{main}
}

\newpage
\twocolumn[\centering\Large\textbf{Supplementary Material}\\\vspace{1.0em}]
\appendix
\input{Chapters/X_suppl}

\end{document}

%% file: Chapters/0_Abstract.tex
\begin{abstract}
Segmentation is central to clinical diagnosis and monitoring, yet the reliability of modern foundation models in medical imaging still depends on the availability of precise prompts. The Segment Anything Model (SAM) offers powerful zero-shot capabilities, although it collapses under the weak, generic, and noisy prompts that dominate real clinical workflows. In practice, annotations such as centerline points are coarse and ambiguous, often drifting across neighboring anatomy and misguiding SAM toward inconsistent or incomplete masks. We introduce \textbf{SPD}, a \textbf{S}aliency-Guided \textbf{P}rompt \textbf{D}istillation framework that converts these unreliable cues into robust guidance. SPD first learns data-driven anatomical priors through a lightweight saliency head to obtain confident localization maps. These priors then drive Contextual Prompt Distillation, which validates and enriches noisy prompts using cues from anatomically adjacent slices, producing a consensus prompt set that matches the behavior of expert reasoning. A Pairwise Slice Consistency objective further enforces local anatomical coherence during segmentation. Experiments on four challenging MRI and CT benchmarks demonstrate that SPD consistently outperforms existing SAM adaptations and supervised baselines, delivering large gains in both region-based and boundary-based metrics. SPD provides a practical and principled path toward reliable foundation model deployment in clinical environments where only imperfect prompts are available.

\end{abstract}

%% file: Chapters/1_Introduction.tex
\section{Introduction}

Precise segmentation of anatomical structures is essential for numerous clinical applications, including quantitative assessment and treatment planning~\cite{bertmi2018, alac}. Although deep learning models such as U-Net~\cite{unet} and its variants~\cite{transunet, nnunet, unetmanba} have achieved impressive success in this task, their performance heavily relies on large-scale annotated datasets~\cite{LimitReview, large-semi}. In clinical practice, however, acquiring such datasets is challenging due to the high cost of expert annotations and stringent privacy regulations~\cite{expansive_medical1}. This scarcity of annotated data significantly limits real-world performance, thereby constraining clinical applicability~\cite{scribble}. While this motivates the use of large-scale pre-trained models to alleviate annotation demands~\cite{scarcefonda}, 
adapting such foundation models to medical imaging introduces a different bottleneck, the reliability of clinical prompts. Crucially, even when pixel-level ground-truth masks are available for supervised training, inference-time performance remains sensitive to the quality of the input prompts. In clinical practice, such prompts are often noisy and imprecise, creating the central challenge tackled in this work.

\begin{figure}[t]
    \centering
    \includegraphics[width=0.48\linewidth]{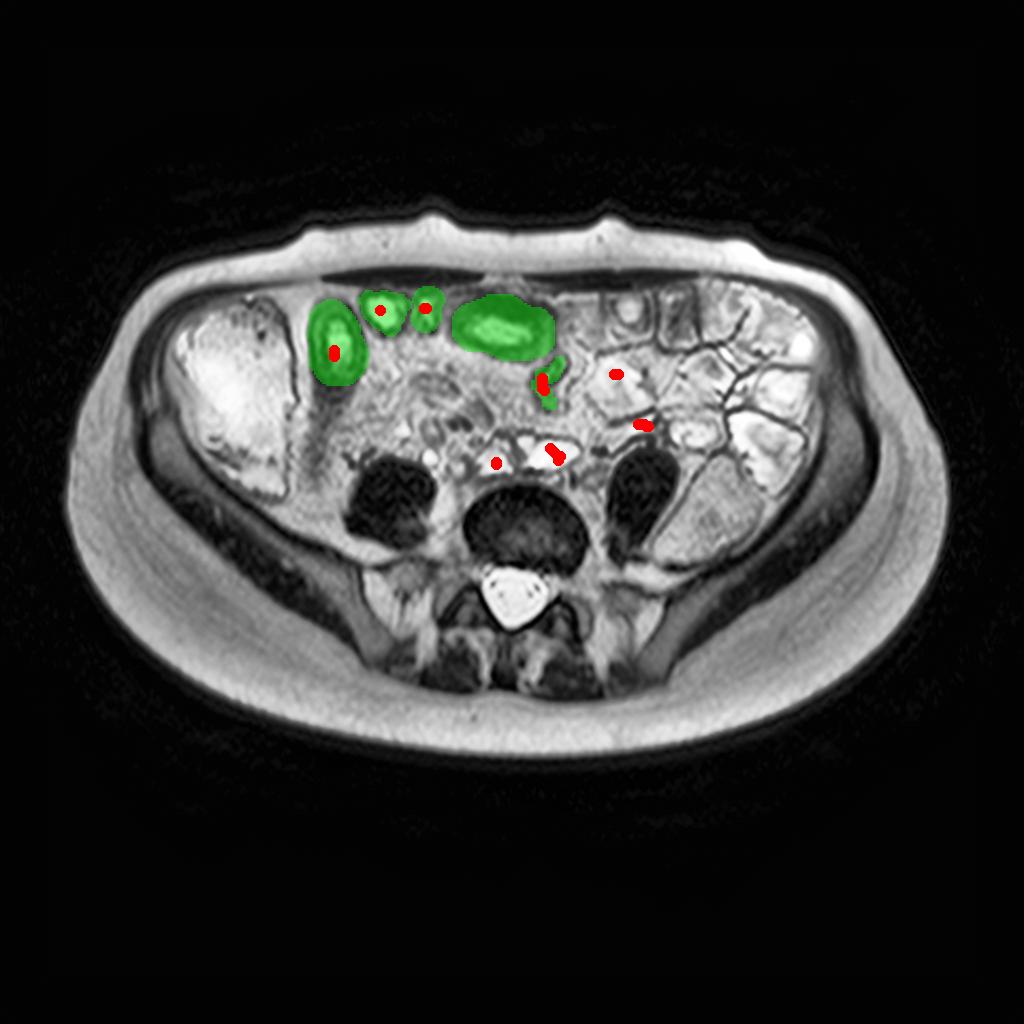}
    \hspace{4pt}
    \includegraphics[width=0.48\linewidth]{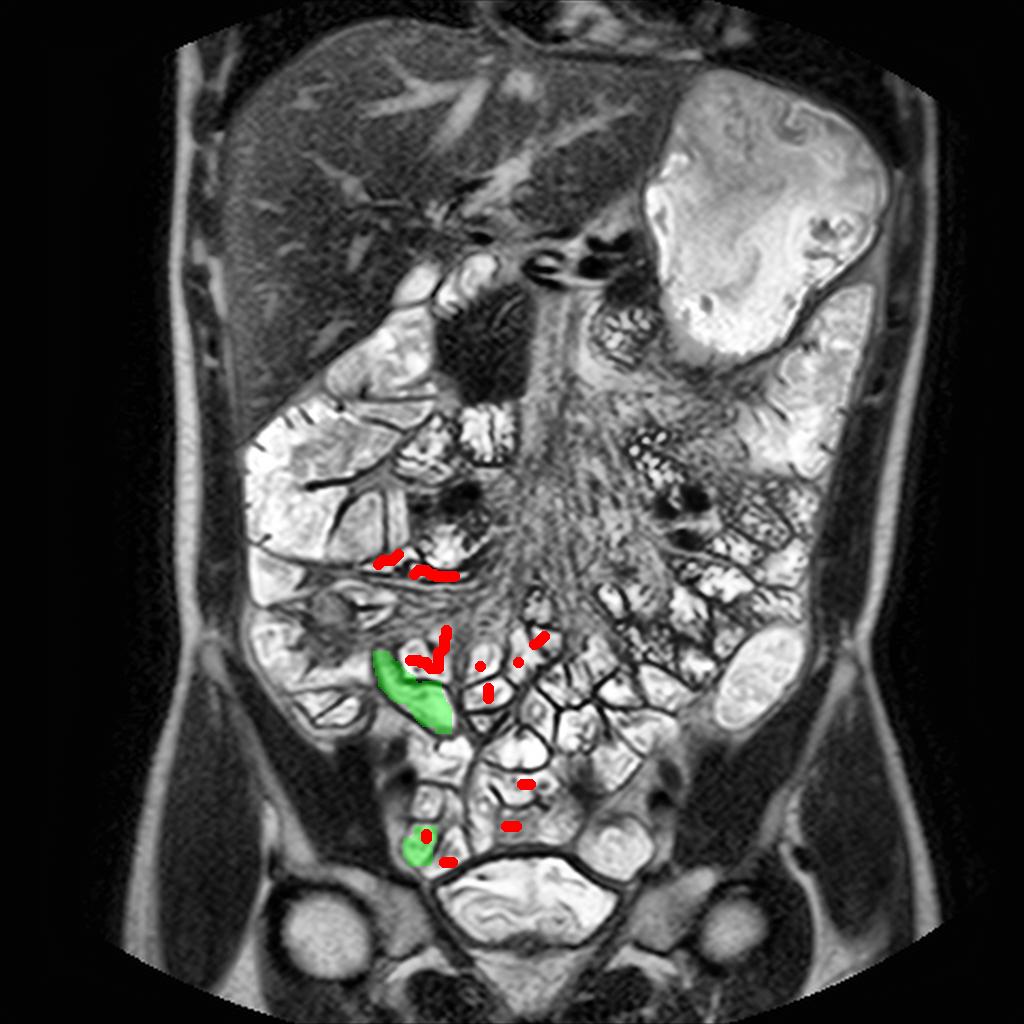}
    \caption{ \textbf{Noisy Prompts.} Radiologists’ centerline annotations (red points) and ground-truth masks (green semi-transparent regions) on (left) an axial and (right) a coronal abdominal MRI view of the terminal ileum (TI). Prompts are derived from real clinical annotations, which are not TI-specific: some TI regions remain unannotated despite being part of the green ground-truth masks, while several annotations extend into adjacent segments such as the mid-ileum or proximal colon. Such noisy prompts mislead SAM during adaptation.}
    \label{fig1}
\end{figure}

Recently, the Segment Anything Model (SAM)~\cite{sam} has attracted attention for its remarkable zero-shot segmentation capability. However, its performance depends heavily on the availability of high-quality prompts and exhibits strong sensitivity to them~\cite{crossconsistent}: SAM segments strictly along the prompted regions, and inaccurate prompts can substantially mislead the model~\cite{fan2023stablesegmentmodel}. In natural images, prompts such as points or bounding boxes are readily obtainable~\cite{howtoprompt, howtoprompt2}, enabling SAM to leverage its pre-trained representations effectively. In contrast, acquiring reliable prompts in medical images requires considerable domain expertise, making them both costly and difficult to obtain~\cite{wu2023self}. In clinical practice, task-specific high-quality prompts are rarely available, and only noisy cues such as centerline points or coarse bounding boxes are typically accessible~\cite{cheng2023sam}. Such prompts are often not tailored specifically for downstream segmentation tasks but rather loosely delineate broader anatomical regions, leading to ambiguity and reduced segmentation accuracy. This work therefore focuses on the prompt quality bottleneck in adapting foundation models like SAM to medical imaging.

Taking terminal ileum (TI) segmentation as an illustrative example, accurate TI assessment is essential for Crohn’s disease diagnosis~\cite{Crohn_D}. However, clinical centerline annotations are primarily designed for general bowel evaluation rather than TI-specific delineation. As shown in Fig.~\ref{fig1}, radiologists’ annotations, which are acquired for routine clinical assessment, often extend into adjacent anatomical segments such as the mid-ileum or proximal colon. Meanwhile, certain regions of the terminal ileum may remain unannotated. Due to SAM’s strict adherence to the prompted regions, noisy prompts misguide SAM, leading to performance degradation during adaptation.

To address this noisy prompts challenge, we propose Saliency-Guided Prompt Distillation (SPD), a two-stage framework that emulates the expert’s reasoning process to adapt foundation models like SAM~\cite{sam} under noisy prompts. Radiologists identify target regions based on a comprehensive understanding of anatomical structures~\cite{radiologist}. When a single slice does not provide sufficient information for a confident assessment, they refer to adjacent slices to gather additional supporting evidence~\cite{nextframe1,nextframe2}. Inspired by this workflow, SPD first learns data-driven anatomical priors via a lightweight saliency head, producing reliable saliency maps that serve as initial localization cues. Building upon these priors, the framework then performs Contextual Prompt Distillation (CPD), a process that leverages contextual information from neighboring slices to rigorously validate local prompts and enrich the overall guidance, ultimately formulating a reliable consensus prompt set. To preserve the anatomical coherence, this segmentation process is further guided by a Pairwise Slice Consistency (PSC) objective, which enforces anatomical consistency across adjacent slices and enhances robustness in ambiguous regions.

\noindent\textbf{The main contributions of this work are as follows:}
\begin{itemize}
    \item We identify and address the critical challenge of adapting foundation models such as SAM to medical imaging under noisy prompts, a scenario commonly encountered in clinical practice.
    
    \item We propose SPD, a two-stage framework that emulates the expert’s diagnostic reasoning process. SPD leverages Contextual Prompt Distillation to distill reliable task-relevant consensus prompts from noisy prompts, and enforces anatomical coherence across adjacent slices via a Pairwise Slice Consistency objective.

    \item We conduct extensive experiments on four challenging medical image segmentation datasets, with modalities ranging from MRI to CT, demonstrating that SPD consistently outperforms state-of-the-art approaches, including recent 3D SAM-based methods. These results highlight its potential as a reliable and practical tool for deploying foundation models in real-world clinical settings where only noisy prompts are available.
\end{itemize}

%% file: Chapters/2_Related_Work.tex
\section{Related Work}

\paragraph{State-of-the-Art Medical Image Segmentation.}  

Deep learning architectures like U-Net~\cite{unet} and its variants~\cite{swinunet, attentionunet, nnunet, xu2025mambavesselnet++} have become the de facto standard for medical image segmentation, achieving impressive results with sufficient annotated data. To reduce the dependency on extensive labeling, paradigms such as self-supervised~\cite{models2, yang2023keypoint} and few-shot learning~\cite{catnet, tang2025few, ding2023few} have been introduced. While these approaches have advanced the field, their performance can be limited compared to large-scale foundation models, which benefit from highly robust and generalizable representations learned from massive datasets~\cite{huang2024segment, radford2021learning}. This has led to a growing interest in adapting these powerful foundation models to leverage their capabilities in the medical domain.

\paragraph{Segment Anything Model in Medical Image Segmentation.}  
SAM~\cite{sam} has rapidly become a focal point for this paradigm shift. Adaptations range from full fine-tuning on large-scale medical datasets, as seen in MedSAM \cite{medsam}, to more parameter-efficient approaches using adapter modules, such as Medical-SAM-Adapter~\cite{medsamadapter} and SAM-Med2D~\cite{sammed2d}. Further explorations include 3D-aware adaptations like MA-SAM~\cite{masam} and automatic prompt generation in AutoSAM~\cite{autosam}. Beyond these adaptations, SAM2~\cite{sam2} extends the foundation model to video segmentation through a memory-based architecture, yet it continues to rely on accurate prompts to initialize and maintain segmentation. While these methods differ architecturally, they all assume access to precise, task-specific prompts, often simulated from ground truth for validation~\cite{medsam}. In medical imaging, however, producing such prompts demands significant domain expertise~\cite{cheng2023sam}, creating a critical gap between practice and real clinical workflows where available prompts are sparse and noisy.

\paragraph{Learning from Noise.}
While our work is broadly related to the field of learning from imperfect supervision, such as learning with noisy labels~\cite{song2022learning,zhang2020characterizing, liu2022adaptive}, it addresses a distinct and fundamentally different challenge. The primary focus of noisy label learning is to develop methods that can robustly train models in the presence of incorrect annotations. Such label noise may arise from random class flips or from more structured, instance-dependent corruption~\cite{karimi2020deep}. In SAM-based segmentation, SAM-Refiner~\cite{refiner} aims to improve robustness by refining SAM outputs using externally provided coarse masks. However, this line of work still relies on supervision in the form of imperfect masks and does not address the unreliability of prompts themselves.
In contrast, our setting is driven by noisy prompts, a distinct and clinically motivated challenge. These prompts are not intrinsically noisy; instead, because precise annotations are costly, task-specific prompts are rarely obtainable. Practitioners must therefore rely on generic anatomical cues, such as centerline points that become noisy when applied to a specific downstream segmentation task. Our SPD framework is designed explicitly for this scenario, providing the first solution tailored to robustness under noisy prompts.

%% file: Chapters/3_Method.tex
\section{Method}

\begin{figure*}[t] 
    \centering
    \includegraphics[width=1\linewidth]{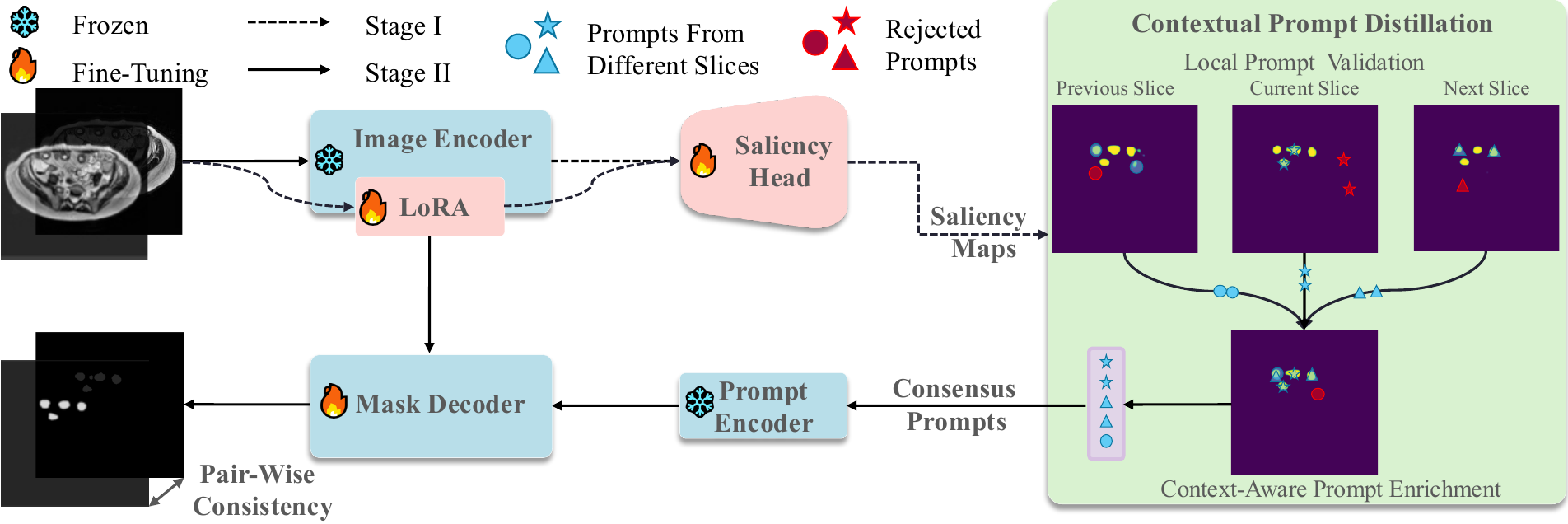}
    \caption{\textbf{Overview of our SPD framework.}
\textit{Stage I (Anatomical Prior Learning):} A lightweight saliency head is trained alongside LoRA-adapted encoder features to generate a high-confidence saliency map, which serves as a reliable anatomical prior.
\textit{Stage II (Prompt-Guided Segmentation):} Saliency map guides the \textit{Contextual Prompt Distillation} module to filter local prompts and integrate consistent cues from neighboring slices (indicated by different shapes), forming a consensus prompt set. The final segmentation is produced by the SAM decoder using these consensus prompts, while a \textit{Pairwise Slice Consistency} objective enforces spatial coherence between adjacent slice predictions.
}
    \label{fig:main_figure}
\end{figure*}

\subsection{Overview}
Our goal is to adapt prompt-based foundation models such as SAM~\cite{sam} to medical image segmentation under noisy prompts. 
Inspired by the diagnostic process of radiologists~\cite{nextframe1, nextframe2, radiologist}, we propose Saliency-Guided Prompt Distillation (SPD), a two-stage framework to address this issue (Fig.~\ref{fig:main_figure}). 
The first stage, \textit{Anatomical Prior Learning}, acquires data-driven anatomical priors and generates saliency maps to provide reliable initial localization cues.
The second stage, \textit{Prompt-Guided Segmentation}, distills consensus prompts for the current slice by leveraging contextual saliency and prompts from adjacent slices, with a pairwise slice consistency objective that ensures anatomical coherence across slices.

We emphasize that SPD follows the standard supervised adaptation paradigm: ground-truth masks are available during training, and the term ``noisy prompts'' refers exclusively to the quality of input prompts at inference time, not to a weakly supervised setting. SPD focuses on robust downstream adaptation rather than preserving SAM's zero-shot behavior; adaptation is implemented via LoRA~\cite{lora}, and disabling the LoRA adapters fully restores the original SAM inference pipeline.

\subsection{Saliency-Guided Anatomical Prior Learning}
\label{sec:saliency_learning}
Radiologists typically form a holistic understanding of anatomical structures before making slice-level decisions, even when noisy cues are provided. Inspired by this observation, we introduce a \textit{lightweight saliency head} to learn data-driven anatomical priors as an initial, reliable cue for subsequent prompt distillation.

The saliency head is a lightweight decoder network that takes features from the LoRA-adapted~\cite{lora} SAM encoder, enabling medical domain adaptation while preserving pre-trained knowledge. Although this module is trained with supervision from the available ground-truth masks, its goal is not to solve the final segmentation task directly. In clinical scenarios, pixel-wise annotations are extremely limited, and a purely supervised segmentation network trained on these scarce labels alone cannot achieve satisfactory performance. We leverage the limited ground-truth masks to learn coarse but reliable anatomical saliency maps that capture where the target structure is plausibly located.

Given an input slice $I_t$, the saliency head $f_{\text{sal}}$ predicts a saliency map $S_t$ highlighting anatomically relevant regions:
\[
S_t = f_{\text{sal}}(I_t;\,\Theta_{\text{sal}}), \quad \Theta_{\text{sal}} = \{\theta_{\text{sal}}, \theta_{\text{lora}}\}
\]
where $\theta_{\text{sal}}$ and $\theta_{\text{lora}}$ denote the parameters of the saliency head and the LoRA adapters, respectively. 
The saliency head is trained under full supervision using ground-truth masks. 
For a training batch of slices, denoted by $\mathcal{B}$, the overall training objective is the average of weighted Dice and Focal Losses computed over all slices in the batch. 
Let $S_t$ and $G_t$ be the predicted saliency map and ground-truth mask for a slice $t \in \mathcal{B}$, respectively. The total loss is then defined as:
\begin{equation}
\mathcal{L}_{\text{sal}} = \frac{1}{|\mathcal{B}|} \sum_{t \in \mathcal{B}} \left(
\lambda_{\text{dice}} \mathcal{L}_{\text{dice}}(S_t, G_t) + 
\lambda_{\text{focal}} \mathcal{L}_{\text{focal}}(S_t, G_t)
\right).
\label{eq:saliency_loss} 
\end{equation}

The Dice Loss is defined as:
\begin{equation}
\mathcal{L}_{\text{dice}}(S_t, G_t) = 1 - \frac{2\sum_{i} S_{t,i} \cdot G_{t,i} + \epsilon}{\sum_{i} S_{t,i} + \sum_{i} G_{t,i} + \epsilon},
\label{eq:dice_loss}
\end{equation}

where $S_{t,i}$ and $G_{t,i}$ denote the predicted probability and ground-truth label at pixel $i$ in slice $t$, and $\epsilon$ is a small constant for numerical stability.

We employ Focal Loss to mitigate class imbalance by reducing the influence of easy background pixels and emphasizing challenging foreground regions during training. 
The Focal Loss is formulated as:
\begin{equation}
\begin{split}
\mathcal{L}_{\text{focal}}(S_t, G_t)
= -\frac{1}{N} \sum_{i} &
\Big[
    G_{t,i}(1-S_{t,i})^{\gamma}\log(S_{t,i}+\epsilon) \\
    &+ (1-G_{t,i}) S_{t,i}^{\gamma}\log(1-S_{t,i}+\epsilon)
\Big],
\end{split}
\label{eq:focal_loss}
\end{equation}
where $\gamma = 2.0$.

\subsection{Contextual Prompt Distillation}
The saliency maps generated in the first stage for all slices serve as reliable, data-driven anatomical priors. In this second stage, our CPD module leverages these priors to address the challenge of noisy prompts. CPD constructs a reliable consensus prompt set for the current slice through a sequence of three key elements: Local Prompt Validation, Context-Aware Prompt Enrichment, and Consensus Set Formulation. 

Formally, our method processes each slice $I_t$ individually. Given a current slice $I_t$ with its noisy prompt set $P_t = \{p_t^1, p_t^2, \dots, p_t^{m_t}\}$, where $m_t$ denotes the number of available prompts for this slice, SPD also references $2n$ neighboring slices $\{I_{t-i}, I_{t+i} \mid i=1,\dots,n\}$ with their corresponding prompt sets $\mathcal{P}_t^{ctx} = \{P_{t-i}, P_{t+i} \mid i=1,\dots,n\}$. 
Saliency maps are generated for all these slices, denoted as $S_t$ for the current slice and $\mathcal{S}_t^{ctx} = \{S_{t-i}, S_{t+i} \mid i=1,\dots,n\}$ for the neighboring slices. 
The consensus prompts $P_t^*$ are obtained through our SPD strategy based on $P_t$, $\mathcal{P}_t^{ctx}$, and $S_t$, and are then used to guide the prediction of the segmentation mask $M_t$ for $I_t$.

\subsubsection{Local Prompt Validation}
The CPD process begins by validating the initial prompt set, $P_t$, which consists of the noisy prompts provided by clinicians. 
To filter out prompts that are anatomically inconsistent with the target structure, we leverage the saliency map $S_t$, which is generated during our \textit{Saliency-Guided Anatomical Prior Learning} stage (Section~\ref{sec:saliency_learning}) as a reliable spatial reference. 
By scrutinizing each prompt against this map, we retain only those located in high-saliency regions, effectively suppressing misleading cues. 
This validation step produces a trustworthy set of local prompts, $\tilde{P}_t^{\text{local}}$:
\begin{equation}
\tilde{P}_t^{\text{local}} = \{p \in P_t \mid S_t(p) > \tau\},
\label{eq:local_prompts}
\end{equation}
where $\tau$ is a predefined saliency threshold.

\subsubsection{Context-Aware Prompt Enrichment}
To augment the prompt set, particularly when local prompts are sparse or insufficient, we employ a mechanism for context-aware enrichment. 
This involves a rigorous dual-validation process to incorporate validated prompts from neighboring slices. 
First, we define the set of neighboring slice indices as $\mathcal{N}_t = \{t \pm i \mid i=1,\dots,n\}$. 
A candidate set, $\mathcal{P}_{\text{cand}}$, is then formed by collecting prompts from these neighbors that are salient \textit{within their own contexts}:
\begin{equation}
\mathcal{P}_{\text{cand}} = \bigcup_{j \in \mathcal{N}_t} \{p \in P_j \mid S_j(p) > \tau\}.
\label{eq:cand_prompts}
\end{equation}

These candidates are then cross-validated against the \textit{current slice's} saliency map $S_t$. Only prompts that pass this second validation are used to enrich the set, forming the contextually-validated prompts $\tilde{P}_t^{\text{ctx}}$:
\begin{equation}
\tilde{P}_t^{\text{ctx}} = \{p \in \mathcal{P}_{\text{cand}} \mid S_t(p) > \tau\}.
\label{eq:ctx_prompts}
\end{equation}

\subsubsection{Consensus Prompt Set Formulation}
In the final stage, the validated local prompts and the enriched contextual prompts are integrated into a single, comprehensive consensus set, $P_t^*$:
\begin{equation}
P_t^* = \tilde{P}_t^{\text{local}} \cup \tilde{P}_t^{\text{ctx}}.
\label{eq:consensus_prompts}
\end{equation}
This high-quality, distilled prompt set is then used to guide the adapted SAM toward a precise and robust segmentation.

\begin{table*}[ht]
\centering
\caption{Performance comparison on TI, Scar, FUMPE, and KiTS datasets. Conventional supervised methods and SAM-based methods are presented separately. \textbf{\textsuperscript{*}} denotes a statistically significant improvement over all the comparison methods (\emph{p}~$<$~0.05), validated using the Wilcoxon signed-rank test.}
\label{tab:main_table}
\resizebox{\textwidth}{!}{%
\begin{tabular}{lcccccccccccccccc}
\toprule
& \multicolumn{4}{c}{TI} & \multicolumn{4}{c}{Scar} & \multicolumn{4}{c}{FUMPE} & \multicolumn{4}{c}{KiTS} \\
\cmidrule(lr){2-5}\cmidrule(lr){6-9}\cmidrule(lr){10-13}\cmidrule(lr){14-17}
Method & DSC$\uparrow$ & IOU$\uparrow$ & HD95$\downarrow$ & ASD$\downarrow$ &
DSC$\uparrow$ & IOU$\uparrow$ & HD95$\downarrow$ & ASD$\downarrow$ &
DSC$\uparrow$ & IOU$\uparrow$ & HD95$\downarrow$ & ASD$\downarrow$ &
DSC$\uparrow$ & IOU$\uparrow$ & HD95$\downarrow$ & ASD$\downarrow$ \\
\midrule
\multicolumn{17}{l}{\textbf{Conventional methods}} \\
UNet & 44.22 & 38.63 & 77.15 & 19.27 & 61.16 & 54.58 & 17.81 & 3.90 & 81.73 & 77.35 & 18.47 & 4.67 & 88.34 & 84.08 & 11.72 & 3.09 \\
UNet++ & 51.82 & 46.72 & 50.57 & 17.87 & 60.28 & 53.84 & 18.69 & 4.31 & 79.72 & 75.42 & 19.04 & 4.47 & 84.18 & 80.45 & 11.04 & 2.48 \\
TransUNet & 52.00 & 47.50 & 33.16 & 11.40 & 55.24 & 48.24 & 19.63 & 4.40 & 71.61 & 66.98 & 22.48 & 4.83 & 67.92 & 58.21 & 24.16 & 7.00 \\
nnUNet & 46.63 & 44.24 & 30.22 & 11.60 & 61.93 & 55.55 & 15.75 & 3.89 & 84.26 & 80.36 & 15.11 & 3.60 & 89.64 & 85.44 & 11.38 & 2.54 \\
\midrule
\multicolumn{17}{l}{\textbf{SAM-based methods}} \\
SAM-Tuning & 62.50 & 55.17 & 37.66 & 9.98 & 70.05 & 62.47 & 21.75 & 4.82 & 68.80 & 68.52 & 93.65 & 54.65 & 88.27 & 84.10 & 11.99 & 2.79 \\
SAM-Refiner & 27.12 & 21.21 & 114.14 & 37.73 & 58.45 & 38.33 & 16.56 & 4.27 & 79.86 & 75.80 & 13.86 & 3.82 & 86.17 & 81.15 & 16.46 & 3.09 \\
MSA & 60.31 & 53.21 & 87.86 & 21.06 & 68.86 & 61.28 & 20.22 & 4.71 & 84.02 & 79.42 & 22.67 & 4.74 & 84.70 & 80.52 & 12.38 & 3.26 \\
MedSAM & 57.26 & 51.36 & 78.97 & 22.88 & 64.33 & 57.67 & 24.75 & 6.53 & 80.05 & 75.99 & 30.07 & 8.50 & 65.20 & 55.96 & 32.11 & 8.71 \\
\textbf{Ours} & \textbf{73.58\textsuperscript{*}} & \textbf{66.22\textsuperscript{*}} & \textbf{23.94\textsuperscript{*}} & \textbf{6.79\textsuperscript{*}} &
\textbf{76.42\textsuperscript{*}} & \textbf{68.62\textsuperscript{*}} & \textbf{12.91\textsuperscript{*}} & \textbf{3.12\textsuperscript{*}} &
\textbf{90.84\textsuperscript{*}} & \textbf{86.75\textsuperscript{*}} & \textbf{12.06\textsuperscript{*}} & \textbf{2.99\textsuperscript{*}} &
\textbf{91.16} & \textbf{86.67} & \textbf{10.12} & \textbf{2.14} \\
\bottomrule
\end{tabular}}
\end{table*}

\subsection{Pairwise Slice Consistency}
Accurate segmentation of volumetric medical images requires anatomical continuity across slices, especially in regions with ambiguous or weakly visible boundaries. Radiologists often reason about such regions by referring to immediately adjacent slices rather than enforcing strict global consistency across the entire volume. Similarly, PSC is designed to regularize local anatomical coherence while avoiding overly strong smoothness assumptions.

Instead of enforcing chain-like consistency across all slices, which induces global smoothness and can propagate errors over long distances~\cite{li2024self}, PSC restricts regularization to immediate slice pairs. This localized constraint strengthens robustness in ambiguous regions while preserving natural anatomical variation across the volume.

\noindent\textbf{Pairwise Consistency Objective.}
Given the predicted probability segmentation maps $(M_t, M_{t+1})$ for two neighboring slices $(I_t, I_{t+1})$, PSC enforces spatial coherence by minimizing the dissimilarity between their predictions. We formulate this as a PSC loss, which converts the maximization of the Dice Similarity Coefficient into a minimization problem for the optimizer:
\begin{equation}
\mathcal{L}_{\text{psc}}(M_t, M_{t+1}) = 1 - \frac{2\sum_{i} M_{t,i} \cdot M_{t+1,i}}{\sum_{i} M_{t,i} + \sum_{i} M_{t+1,i}},
\label{eq:psc_loss}
\end{equation}
where the Dice coefficient is computed on the soft probability maps to ensure the loss is differentiable.

\subsection{Overall Training Objective}
SPD is optimized in two distinct, sequential stages.

\textbf{Stage 1: Anatomical Prior Learning.}
We train the saliency head along with the LoRA adapters of the image encoder. The objective is to learn robust anatomical priors from ground-truth masks by minimizing the saliency loss $\mathcal{L}_{\text{sal}}$, as defined in Equation~\ref{eq:saliency_loss}. Upon completion of this stage, the parameters of the saliency head ($\theta_{\text{sal}}$) and the LoRA adapters ($\theta_{\text{lora}}$) are frozen.

\textbf{Stage 2: Prompt-Guided Segmentation.}
With the saliency head and adapted encoder acting as fixed modules, we fine-tune only the SAM mask decoder. The model is trained using the consensus prompts $P_t^*$ distilled by our CPD module. The overall training objective is computed over a mini-batch of slices $\mathcal{B}$. The total loss for the batch averages the combined segmentation and pairwise consistency objectives for each slice in the batch:
\begin{equation}
\mathcal{L}_{\text{total}} = \frac{1}{|\mathcal{B}|} \sum_{t \in \mathcal{B}} \left( \mathcal{L}_{\text{seg}}(M_t, G_t) + \lambda_{\text{psc}} \mathcal{L}_{\text{psc}}(M_t, M_{t+1}) \right).
\label{eq:total_loss}
\end{equation}
 The segmentation loss $\mathcal{L}_{\text{seg}}$ shares the same formulation as $\mathcal{L}_{\text{sal}}$ (Equation~\ref{eq:saliency_loss}).

%% file: Chapters/4_Experiment.tex
\section{Experiment}

\subsection{Datasets}
We evaluated SPD on two private medical MRI segmentation datasets and two public CT Dataset to demonstrate its general applicability. For all experiments, we use a unified experimental setup. All slices were resampled to a uniform resolution of $512 \times 512$. For the TI dataset, we directly use the available clinical annotations centerline points as natural prompts, which reflect real-world annotation patterns. For other datasets lacking such clinical annotations, we simulated noisy prompts to approximate these real-world scenarios. Our simulation strategy extends the approach from MedSAM~\cite{medsam}: starting with a single point sampled from within the ground-truth mask, we add 2-5 additional points sampled from the entire image area. All points were treated as positive prompts, requiring the model to identify the true anatomical structure from sparse and potentially misleading spatial cues. For CPD, we set the saliency threshold $\tau = 0.5$ and the contextual slice number $n = 2$ (i.e., $2n{+}1 = 5$ slices in total) across all datasets. Additional implementation details are available in Appendix~\ref{sec:implement_details}.

\begin{figure*}[t] 
    \centering
    \includegraphics[width=1\linewidth]{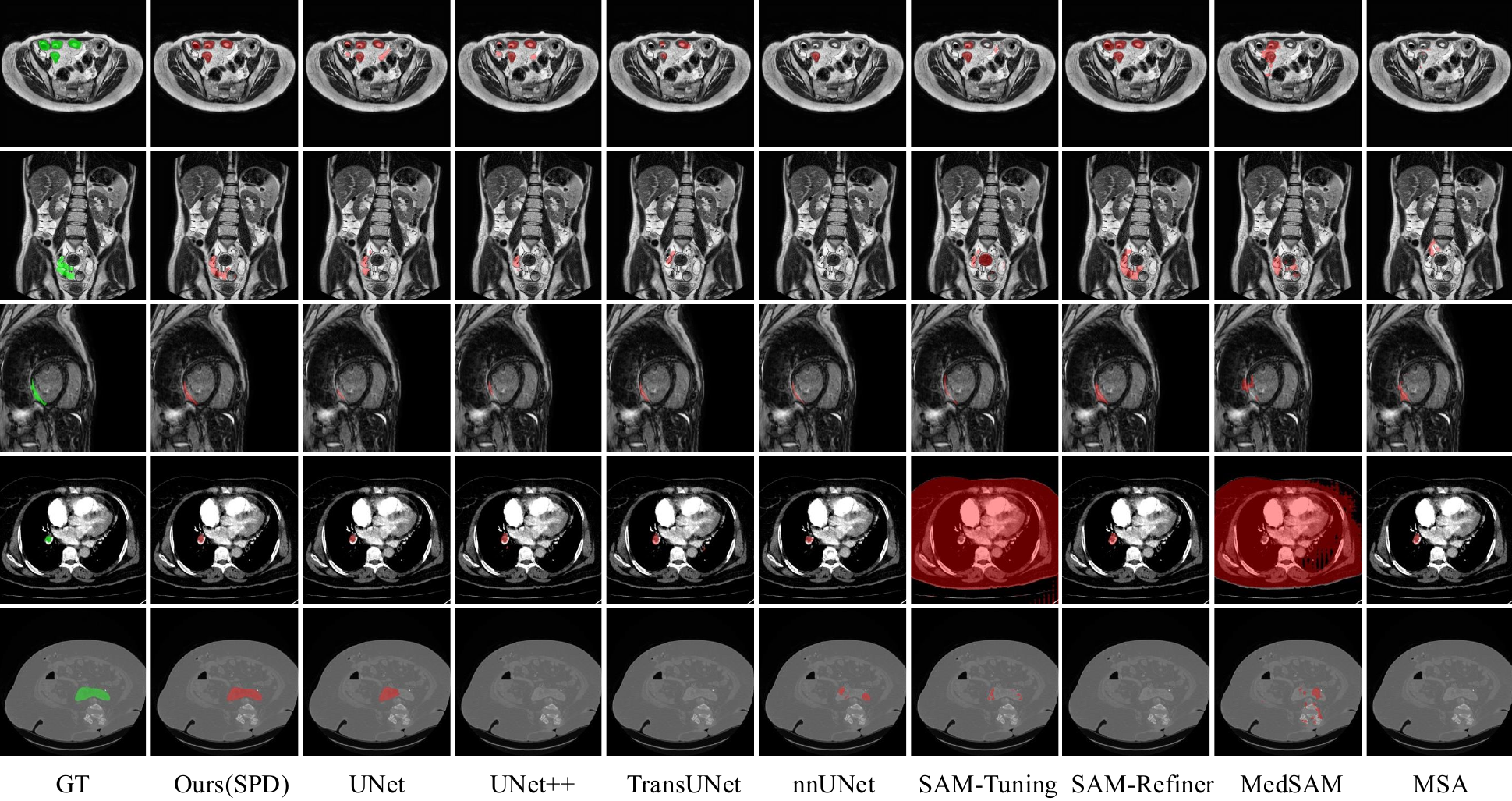}
    \caption{\textbf{Qualitative comparison.} 
       Each column corresponds to a different method. The ground truth is overlaid as a semi-transparent green mask, while predictions are outlined in red. The top two rows present challenging axial and coronal views from the TI dataset, and the bottom three rows show results on the Scar, FUMPE and KiTS datasets.}
    \label{fig:visualization}
\end{figure*}

\noindent\textbf{Crohn's Disease Terminal Ileum Segmentation Dataset (TI)}. We utilize a T2-weighted MRI dataset collected from a collaborating hospital, comprising 47 subjects.
Each subject was associated with axial and coronal T2 images. The number of slices per volume ranged from 34 to 88, and the original in-plane resolutions varied.
Each scan covers a field of view of 375 mm, with inter-slice spacing between 3.0 mm and 3.5 mm. Experienced radiologists manually labeled pixel-wise annotations of the terminal ileum region. In addition, centerline coordinates are provided to trace anatomical structures from the mid ileum through the terminal ileum to the proximal colon.

\noindent\textbf{Heart Scar Dataset (Scar)}.
This dataset, provided by Nottingham University Hospitals NHS Trust focuses on segmentation of scar tissue within the myocardium using Late Gadolinium Enhancement (LGE) Cardiac Magnetic Resonance (CMR) imaging. It comprises scans from 400 patients, with each subject containing raw LGE images along with corresponding ground truth annotations manually delineated by clinical experts using ITK-Snap. The LGE images present short-axis cross-sectional views of the left and right ventricles. Each subject includes 9-15 image slices. All scans are resampled to have a consistent inter-slice (z-axis) spacing of 10mm.

\noindent\textbf{Ferdowsi University of Mashhad Pulmonary Embolism Dataset (FUMPE)}. \cite{FUMPE}
This dataset is publicly available and specifically designed for the segmentation of pulmonary embolism (PE) lesions in computed tomography angiography (CTA) scans. The dataset comprises 35 patients, each represented by a series of 139 to 475 axial 2D slices, providing detailed volumetric imaging data.

\noindent\textbf{The Kidney and Kidney Tumor Segmentation Challenge 2023 (KiTS)}.~\cite{kits}
KiTS provides 599 contrast-enhanced abdominal CT scans with voxel-level annotations covering kidneys, renal tumors, and renal cysts. The dataset spans multiple acquisition phases and exhibits substantial anatomical and intensity variability, making it a challenging benchmark for abdominal organ and lesion segmentation.

\subsection{Evaluation Metrics}
We evaluate segmentation performance using four standard metrics. Specifically, we report the average Dice Similarity Coefficient (Avg DSC) and average Jaccard Index (Avg IoU), computed as the mean of per-image scores across the test set, to assess region-level overlap between predicted and ground-truth masks. For boundary-level accuracy, we use the 95\% Hausdorff Distance (HD95) to capture the largest boundary deviation while remaining robust to small outliers, and the Average Surface Distance (ASD) to quantify the mean surface discrepancy. More details are provided in Appendix~\ref{sec:eval_metrics_details}.

\subsection{Overall Results}
\noindent\textbf{Comparison with State-of-the-Art Methods.} 
We evaluate our SPD framework against established segmentation methods and recent SAM-based approaches across four challenging medical imaging datasets. Table~\ref{tab:main_table} demonstrates that our method achieves consistent and significant improvements over all baselines.

Conventional supervised methods, including CNN-based architectures (UNet~\cite{unet}, UNet++~\cite{unetplusplus}, nnUNet~\cite{nnunet}) and transformer-based models (TransUNet~\cite{transunet}), exhibit limited performance on our medical imaging benchmarks. Despite architectural innovations, such as automated hyperparameter optimization in nnUNet~\cite{nnunet} and global attention mechanisms in TransUNet~\cite{transunet}, these methods struggle with the inherent challenges of medical imaging: scarce annotations, high inter-patient variability, and complex anatomical structures in both MRI and CT modalities.

\begin{figure}[t] 
\centering
\includegraphics[width=\linewidth]{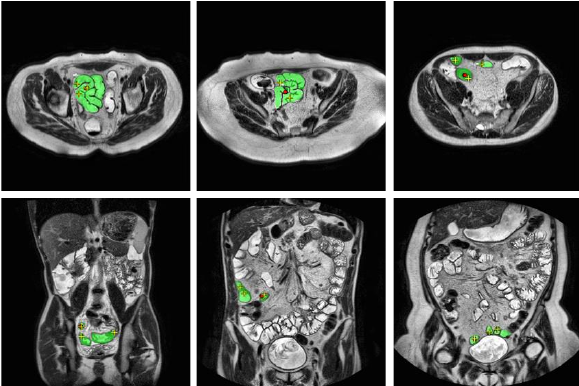}
\caption{\textbf{Visualization of consensus prompts on the TI dataset.}  Green masks denote ground truth annotations. Red dots are prompts selected from the current slice, and yellow crosses are prompts propagated from neighboring slices. These together constitute the final consensus prompt set.}
\label{fig:supp_prompt_selection}
\end{figure}

SAM-based adaptation methods, including SAM-Tuning~\cite{sam}, MedSAM~\cite{medsam}, Medical-SAM-Adapter (MSA)~\cite{medsamadapter}, demonstrate improved performance but remain fundamentally limited by their dependence on high-quality prompts. When faced with the noisy prompts characteristic of the real-world data, such as centerline annotations in the TI dataset, these methods often struggle to produce reliable segmentations, leading to degraded performance. SAM-Refiner~\cite{refiner} refines SAM predictions using externally provided coarse masks, but it still cannot effectively handle noisy prompts, as its performance depends on the quality of those masks rather than addressing prompt noise directly. Moreover, it cannot determine meaningful refinements when the provided mask is empty, further limiting its robustness in real clinical scenarios.

SPD directly addresses the prompt sensitivity bottleneck. By distilling reliable guidance from noisy prompts via our saliency-guided contextual framework, SPD demonstrates significant performance improvements over state-of-the-art methods across all evaluated datasets. On the TI dataset, SPD achieves 11.08\% increase in DSC and a remarkable 6.28 reduction in HD95 compared to the best existing approach. For the Scar dataset, SPD delivers a 6.37\% improvement in DSC and 8.84 reduction in HD95. Similarly, for the FUMPE dataset, our method surpasses the previous best model with improvements of 6.58\% in DSC and 3.05 in HD95. For the KiTS dataset, SPD achieves the highest accuracy, improving DSC by 1.52\% and reducing HD95 by 1.26 compared to the best competing method. These substantial and consistent gains across multiple datasets validate the effectiveness and robustness of our proposed approach in real-world clinical settings.

\noindent\textbf{Comparison with 3D Medical Segmentation Methods.}
We further compare SPD against recent 3D SAM-based methods, SAM2~\cite{sam2} and MedSAM2~\cite{medsam2}, on the TI dataset. As shown in Fig.~\ref{fig:sam2_3d}, SPD consistently outperforms both SAM2 and MedSAM2 under Average Dice and Volumetric Dice metrics. We additionally report Volumetric Dice scores across all four datasets in Table~\ref{tab:vol_dice}, where SPD achieves consistent improvements over the best prior baselines. Full details are in Appendix~\ref{sec:3d_comparison}.

\begin{figure}[t]
    \centering
    \includegraphics[width=\linewidth]{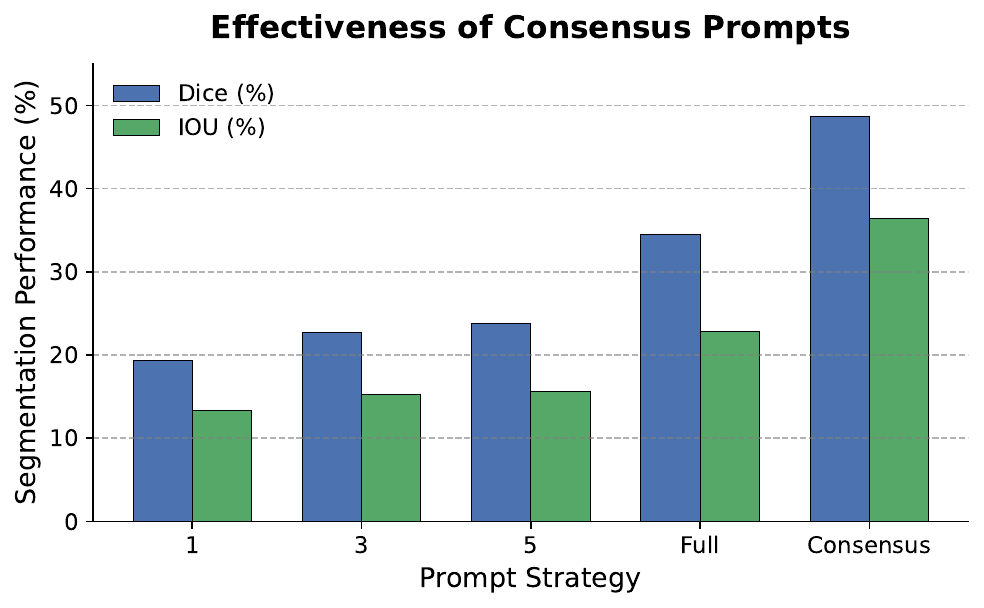}
    \caption{\textbf{Zero-shot performance of a frozen SAM model guided by different prompt sources. }
    ``1'', ``3'', and ``5'' denote the number of randomly sampled points from the original centerline annotations, 
    ``Full'' indicates using all centerline points, and ``Consensus'' refers to our proposed consensus prompts. }
    \label{fig:prompt_effectiveness}
\end{figure}

\noindent\textbf{Qualitative Results.}
Fig.~\ref{fig:visualization} provides qualitative comparisons across all datasets. Traditional supervised methods either miss target regions or incorrectly segment non-target areas. SAM-based baselines are severely misled by noisy prompts; in particular, SAM-Tuning and MedSAM tend to over-segment due to prompt-sensitivity, especially on the FUMPE dataset. In contrast, SPD effectively alleviates this issue by distilling reliable guidance from imperfect prompts, yielding substantially more accurate and anatomically consistent segmentations.

\noindent\textbf{Consensus Prompts Visualization.}
Figure~\ref{fig:supp_prompt_selection} visualizes the consensus prompts produced by our CPD module. Notably, prompts propagated from adjacent slices (yellow crosses) are consistently positioned within the ground-truth structures of the current slice, confirming the effectiveness of our cross-slice enrichment strategy in augmenting sparse local prompts.

\begin{table}[t]
\centering
\caption{\textbf{Ablation study on the TI dataset.} ``Local'' refers to using only current slice prompts. ``CPD'' refers to our CPD strategy. ``PSC'' denotes Pairwise Slice Consistency.}
\label{tab:ablation}
\resizebox{\columnwidth}{!}{
\begin{tabular}{l c c c c c}
\toprule
Method & Local & CPD & PSC & DSC (\%) $\uparrow$ & IoU (\%) $\uparrow$ \\
\midrule
Baseline       &  &  &  & 62.50 & 55.17 \\
Local          & \checkmark &  &  & 66.95 & 60.01 \\
w/o PSC        & \checkmark & \checkmark &  & 70.32 & 63.20 \\
Ours           & \checkmark & \checkmark & \checkmark & \textbf{73.58} & \textbf{66.22} \\
\bottomrule
\end{tabular}
}
\end{table}

\subsection{Ablation Study}
\noindent\textbf{Effectiveness of Consensus Prompts.} 
To evaluate the effectiveness of our proposed consensus prompts, we compare the zero-shot segmentation performance of a frozen SAM model on the TI dataset under different prompt sources. As shown in Fig.~\ref{fig:prompt_effectiveness}, using a few randomly sampled points from the original noisy centerline annotations yields suboptimal results, and even utilizing all centerline points only marginally improves performance due to their weak anatomical relevance. In contrast, our consensus prompts provide significantly more reliable guidance, achieving a substantial improvement of 14.2\% DSC and 13.6\% IOU over the full original centerline points. These results demonstrate that accurately consolidating noisy clinical annotations into consensus prompts can effectively unlock the potential of SAM in medical imaging.

\noindent\textbf{Ablation Study on core components.}
The ablation results in Table~\ref{tab:ablation} verify the effectiveness of each proposed component. Starting from the \textit{Baseline}, which corresponds to SAM with initial prompt tuning, introducing local prompts alone yields a noticeable performance gain, improving the DSC from 62.50\% to 66.95\%. This confirms the benefit of incorporating spatial guidance. Building upon this, our CPD module (\textit{w/o PSC}) brings a substantial boost of over 3\% in DSC, reaching 70.32\%. This highlights CPD's ability to distill informative prompts by integrating contextual cues from neighboring slices. Equipping the model with the PSC objective (\textit{Ours}) further elevates the DSC to 73.58\% and IoU to 66.22\%, indicating that enforcing inter-slice coherence enhances boundary precision and segmentation stability. We also investigate the effect of the contextual slice number n in CPD, and detailed results are provided in Appendix~\ref{sec:slice_num}.

\begin{figure}[t]
    \centering
    \includegraphics[width=\linewidth]{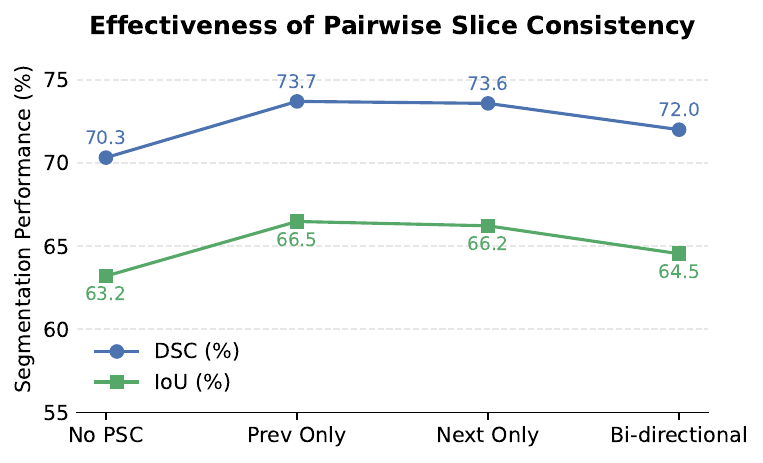}
    \caption{
    \textbf{Impact of guidance source in PSC on segmentation performance.} 
    “No PSC” omits consistency enforcement. “Prev Only” and “Next Only” enforce anatomical coherence with the preceding or succeeding slice, respectively. “Bi-directional” integrates both directions. 
    }
    \label{fig:psc_curve}
\end{figure}

\paragraph{Hyperparameter Sensitivity Analysis.}
We further investigate the impact of guidance source in the Pairwise Slice Consistency mechanism by varying the relative positioning of adjacent slices used for enforcing anatomical coherence. As illustrated in Fig.~\ref{fig:psc_curve}, incorporating consistency with either the preceding slice (``Prev Only'') or the succeeding slice (``Next Only'') significantly improves performance compared to the variant without PSC (``No PSC'').  In contrast, enforcing consistency with both neighboring slices simultaneously (``Bi-directional'') does not lead to additional gains and instead results in a marginal decline in performance. The degradation may arise from conflicting contextual cues: the current slice is forced to align with both preceding and succeeding slices, which can be problematic when they differ substantially. We further assess the robustness of our saliency-guided prompt distillation process with respect to the threshold parameter $\tau$. See Appendix~\ref{sec:Threshold} for detailed distribution analysis.

We further analyze the dependence on Stage~I quality and the computational overhead of SPD in Appendices~\ref{sec:stage1_dependence} and~\ref{sec:computational_cost}.

%% file: Chapters/5_Conclusion.tex
\section{Conclusion}
We presented SPD for adapting foundation models to medical imaging under noisy clinical prompts. By combining anatomical prior learning with contextual prompt distillation, SPD delivers stronger performance than existing methods across diverse MRI and CT datasets. These results support robust prompt handling for reliable clinical deployment in real-world medical settings.

%% file: Chapters/X_suppl.tex
\section{Implementation Details}
\label{sec:implement_details}
We implement SPD using the PyTorch framework and conduct all experiments on 8 NVIDIA RTX 4090 GPUs, with a per-GPU batch size of 4. The model is trained using the Adam optimizer with an initial learning rate of $10^{-4}$ and weight decay of $10^{-5}$.
To stabilize training, we use the \texttt{ReduceLROnPlateau} learning rate scheduler. This scheduler reduces the learning rate by a factor of 0.5 if the validation loss does not improve for 4 consecutive epochs, ensuring efficient learning without overfitting.
We resize all images to $512 \times 512$ pixels. Since the default resolution for SAM is $1024 \times 1024$, we use bilinear interpolation to resize the positional encoding maps to the new size of $512 \times 512$. 
The LoRA configuration is set with \( r = 8 \), \( \alpha = 16 \), and a dropout rate of 0.1. The loss function used is a weighted combination of Dice loss, Focal loss, and Pairwise Consistency with weights set to 0.7, 0.3, and 0.1, respectively. 
Data augmentation is performed using random rotation and horizontal flipping. The images are normalized to $[-1,1]$.

For all datasets, we use consistent parameters to ensure fairness in comparison. The threshold for saliency map filtering is set to 0.5, and we consider a total of 5 slices for contextual reasoning.

\begin{figure}[t]
\centering
\includegraphics[width=\linewidth]{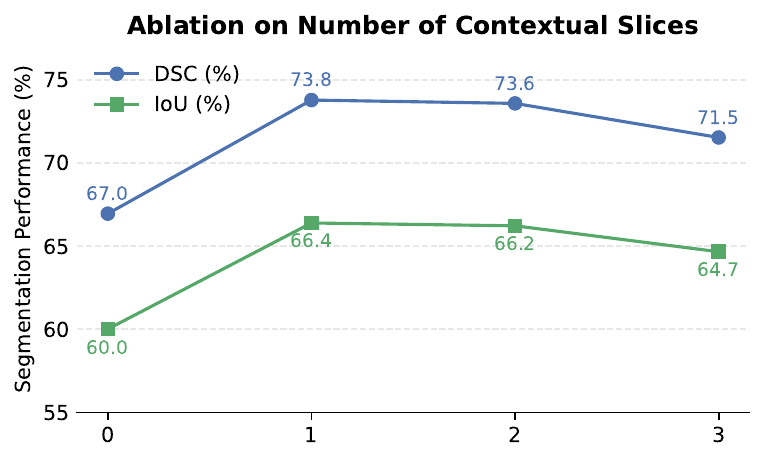}
\caption{\textbf{Ablation on the number of contextual slices \( n \)} in CPD on the TI dataset. }
\label{fig:context_slices_ablation}
\end{figure}
\section{Evaluation Metrics Details}
\label{sec:eval_metrics_details}
This section elaborates on the evaluation metrics used in our experiments, including their definitions and how special cases were handled during computation.

\textbf{Region-based metrics.} The Dice Similarity Coefficient (DSC) and the Jaccard Index (IoU) measure region-level agreement between the predicted and ground-truth segmentation masks. A score of 1.0 indicates perfect overlap, while 0 denotes no overlap. Both metrics are computed per image and averaged over the test set. Higher values indicate better segmentation quality.

\textbf{Boundary-based metrics.} The 95\% Hausdorff Distance (HD95) and Average Surface Distance (ASD) assess boundary alignment between prediction and ground truth. 
HD95 measures the largest distance between boundaries after excluding the top 5\% of outlier points, offering robustness to small errors. Lower values are better. ASD computes the average distance between the predicted and ground-truth surfaces, capturing overall boundary proximity. Again, lower values are preferred.

\textbf{Handling of empty masks.} For cases with empty masks, true negatives are scored as a perfect match, while complete failures receive the worst possible score. Boundary-based metrics such as HD95 and ASD are only computed on samples containing valid segmentation regions.

\begin{figure}[t]
    \centering
    \includegraphics[width=\columnwidth]{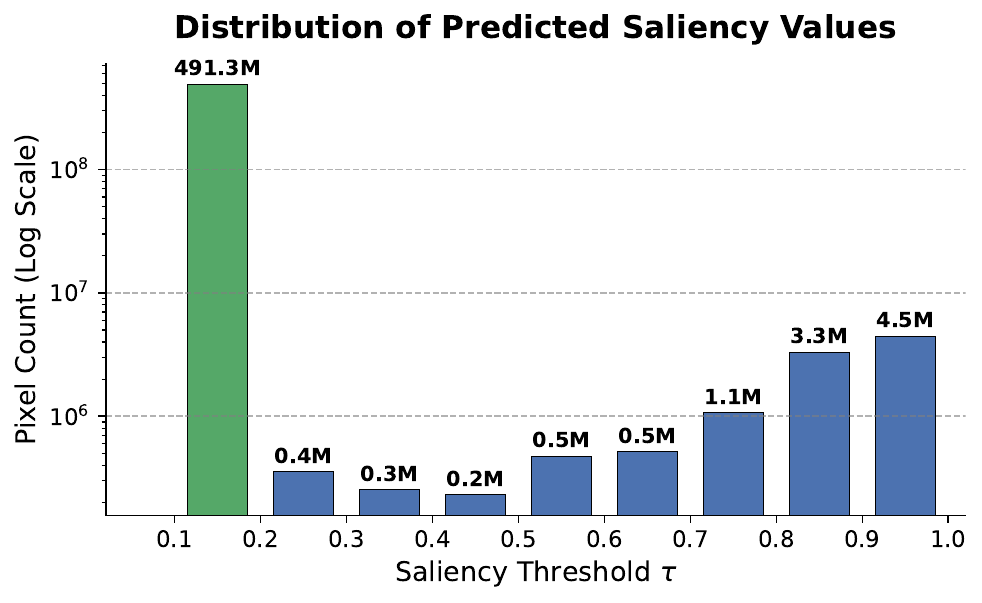} 
\caption{
\textbf{Distribution of predicted saliency values on the TI dataset.}
The histogram is plotted on a logarithmic scale to visualize the full range of pixel counts.
It reveals a highly skewed distribution dominated by a massive peak of high-confidence background predictions (491.3M pixels).
In contrast, all potential foreground signals (values $>$ 0.2) collectively account for less than 2\% of the total pixels, highlighting their extreme sparsity.
This clear separation between confident background and sparse foreground explains the model's robustness to the saliency threshold $\tau$.
}

    \label{fig:saliency_distribution}
\end{figure}

\section{Ablation Study on Contextual Slice Number}
\label{sec:slice_num}
We investigate the impact of the contextual slice number \( n \) in our Contextual Prompt Distillation (CPD) module, where \( n \) denotes the number of adjacent slices on each side of the current slice. This leads to \( 2n+1 \) slices as input for contextual reasoning. We vary \( n \in \{0, 1, 2, 3\} \) and report the results on the TI dataset.

As shown in Figure~\ref{fig:context_slices_ablation}, performance improves significantly when incorporating contextual information, with the Dice score increasing from 66.95\% at \( n=0 \) to 73.78\% at \( n=1 \). Interestingly, while \( n=1 \) achieves the highest average performance, the results remain comparable at \( n=2 \) (73.58\%) and \( n=3 \) (71.53\%), suggesting that our method is not overly sensitive to the choice of \( n \) within a reasonable range.

\section{Robustness to Saliency Threshold \texorpdfstring{$\tau$}{tau}}
\label{sec:Threshold}
To validate the stability of our framework, we analyzed its sensitivity to the saliency threshold $\tau$, a key hyperparameter for our prompt distillation. Specifically, we examined the distribution of all predicted saliency values produced by the trained saliency head across the entire TI dataset, as shown in Figure~\ref{fig:saliency_distribution}.

The histogram reveals a strikingly polarized distribution. A dominant peak appears in the [0.1, 0.2) interval, which alone accounts for 491.3 million pixels—97.88\% of all predictions. In contrast, the pixel counts in other bins are several orders of magnitude smaller: for example, only 0.66\% of the predictions fall into [0.8, 0.9), and 0.89\% into [0.9, 1.0). Intermediate ranges such as [0.3, 0.4), [0.4, 0.5), and [0.5, 0.6) each contribute less than 0.1\%. This leads to a virtual “confidence valley” between the dominant low-saliency predictions and the sparse high-saliency signals.

Such extreme skewness indicates that our model is highly confident in distinguishing background from foreground, rarely assigning ambiguous intermediate values. Importantly, any saliency threshold $\tau$ chosen within this valley (e.g., from 0.3 to 0.8) will yield nearly identical binarization outcomes and validated prompt sets.

This insensitivity to the exact choice of $\tau$ highlights the robustness and hyperparameter stability of our SPD framework, eliminating the need for delicate tuning and simplifying deployment.

\section{Comparison with 3D Medical Segmentation Methods}
\label{sec:3d_comparison}
\begin{figure}[t]
  \centering
  \includegraphics[width=0.96\linewidth]{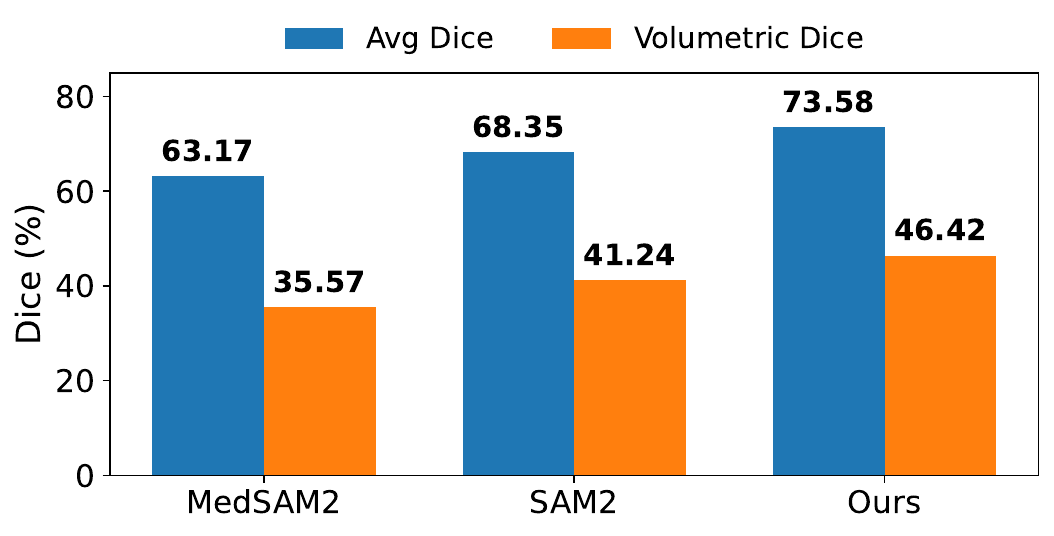}
  \caption{Comparison with recent 3D SAM-based baselines on the Terminal Ileum dataset, evaluated with Average Dice and Volumetric Dice.}
  \label{fig:sam2_3d}
\end{figure}

We compare SPD against recent 3D SAM-based methods, SAM2~\cite{sam2} and MedSAM2~\cite{medsam2}, on the TI dataset. As shown in Figure~\ref{fig:sam2_3d}, SPD consistently outperforms both SAM2 and MedSAM2 under both Average Dice and Volumetric Dice metrics. Note that Volumetric Dice scores are numerically lower than Average Dice due to severe foreground sparsity across slices.

We additionally report Volumetric Dice scores across all four datasets in Table~\ref{tab:vol_dice}. SPD consistently achieves the highest scores, with clear improvements of +7.34\%, +9.08\%, +7.30\%, and +0.95\% over the best prior baseline on TI, Scar, FUMPE, and KiTS, respectively.

\begin{table}[t]
\centering
\small
\setlength{\tabcolsep}{3.2pt}
\renewcommand{\arraystretch}{1.05}
\caption{Volumetric Dice scores. $\Delta$ denotes improvement over the best prior baseline.}
\label{tab:vol_dice}
\resizebox{\linewidth}{!}{%
\begin{tabular}{l|cccccc|c}
\toprule
Dataset &  MedSAM & MSA & SAM-Refiner & UNet & nnU-Net & \textbf{Ours} & $\Delta$ \\
\midrule
TI & 31.54 & 34.83 & 39.08 & 28.64 & 14.36 & \textbf{46.42} & +7.34\% \\
SCAR      & 32.15     & 44.60 & 37.20 & 47.77 & 47.66 & \textbf{56.74} & +9.08\% \\
FUMPE     & 35.33 & 46.14 & 63.09 & 62.43 & 66.18     & \textbf{73.48} & + 7.30\% \\
KiTS      & 75.79     & 91.64 & 90.20 & 91.51 & 93.46 & \textbf{94.41} & +0.95\% \\
\bottomrule
\end{tabular}}
\end{table}

\section{Dependence on Stage~I}
\label{sec:stage1_dependence}
Notably, the saliency head in Stage~I is intentionally coarse, achieving only \textbf{26.95} Dice when evaluated directly against ground-truth masks. It therefore serves as a weak and imperfect anatomical prior rather than an accurate segmentation module. Despite this coarse Stage~I prior, the overall segmentation performance remains strong because the final accuracy is primarily driven by Stage~II prompt-guided segmentation. This demonstrates that our framework does not critically depend on the quality of Stage~I predictions and confirms the robustness of the two-stage design.

\section{Computational Cost}
\label{sec:computational_cost}
SPD introduces only marginal inference overhead. On the Terminal Ileum dataset using a single NVIDIA RTX 4090, SPD runs at 11.49 FPS, compared to 12.68 FPS for standard SAM. The saliency head is a lightweight decoder, and after consensus prompts are constructed, the final segmentation is performed identically to SAM. In practice, saliency maps and consensus prompts can be computed once per volume and cached, making the effective inference latency comparable to existing SAM-based methods.

%% file: main.bib
@String(ICLR = {Int. Conf. Learn. Represent.})

@String(ICLR  = {ICLR})

@ARTICLE{bertmi2018,
  author={Bernard, Olivier and Lalande, Alain and Zotti, Clement and Cervenansky, Frederick and Yang, Xin and Heng, Pheng-Ann and Cetin, Irem and Lekadir, Karim and Camara, Oscar and Gonzalez Ballester, Miguel Angel and Sanroma, Gerard and Napel, Sandy and Petersen, Steffen and Tziritas, Georgios and Grinias, Elias and Khened, Mahendra and Kollerathu, Varghese Alex and Krishnamurthi, Ganapathy and Rohé, Marc-Michel and Pennec, Xavier and Sermesant, Maxime and Isensee, Fabian and Jäger, Paul and Maier-Hein, Klaus H. and Full, Peter M. and Wolf, Ivo and Engelhardt, Sandy and Baumgartner, Christian F. and Koch, Lisa M. and Wolterink, Jelmer M. and Išgum, Ivana and Jang, Yeonggul and Hong, Yoonmi and Patravali, Jay and Jain, Shubham and Humbert, Olivier and Jodoin, Pierre-Marc},
  journal={IEEE Transactions on Medical Imaging}, 
  title={Deep Learning Techniques for Automatic MRI Cardiac Multi-Structures Segmentation and Diagnosis: Is the Problem Solved?}, 
  year={2018},
  volume={37},
  number={11},
  pages={2514-2525},
  doi={10.1109/TMI.2018.2837502}}

@misc{alac,
      title={Deep Learning methods for automatic evaluation of delayed enhancement-MRI. The results of the EMIDEC challenge}, 
      author={Alain Lalande and Zhihao Chen and Thibaut Pommier and Thomas Decourselle and Abdul Qayyum and Michel Salomon and Dominique Ginhac and Youssef Skandarani and Arnaud Boucher and Khawla Brahim and Marleen de Bruijne and Robin Camarasa and Teresa M. Correia and Xue Feng and Kibrom B. Girum and Anja Hennemuth and Markus Huellebrand and Raabid Hussain and Matthias Ivantsits and Jun Ma and Craig Meyer and Rishabh Sharma and Jixi Shi and Nikolaos V. Tsekos and Marta Varela and Xiyue Wang and Sen Yang and Hannu Zhang and Yichi Zhang and Yuncheng Zhou and Xiahai Zhuang and Raphael Couturier and Fabrice Meriaudeau},
      year={2021},
      eprint={2108.04016},
      archivePrefix={arXiv},
      primaryClass={eess.IV},
      url={https://arxiv.org/abs/2108.04016}, 
}

@inproceedings{swinunet,
  title={Swin-unet: Unet-like pure transformer for medical image segmentation},
  author={Cao, Hu and Wang, Yueyue and Chen, Joy and Jiang, Dongsheng and Zhang, Xiaopeng and Tian, Qi and Wang, Manning},
  booktitle={European conference on computer vision},
  pages={205--218},
  year={2022},
  organization={Springer}
}

@article{attentionunet,
  author       = {Ozan Oktay and
                  Jo Schlemper and
                  Lo{\"{\i}}c Le Folgoc and
                  Matthew C. H. Lee and
                  Mattias P. Heinrich and
                  Kazunari Misawa and
                  Kensaku Mori and
                  Steven G. McDonagh and
                  Nils Y. Hammerla and
                  Bernhard Kainz and
                  Ben Glocker and
                  Daniel Rueckert},
  title        = {Attention U-Net: Learning Where to Look for the Pancreas},
  journal      = {CoRR},
  volume       = {abs/1804.03999},
  year         = {2018},
  url          = {http://arxiv.org/abs/1804.03999},
  eprinttype    = {arXiv},
  eprint       = {1804.03999},
  bibsource    = {dblp computer science bibliography, https://dblp.org}
}

@misc{unetplusplus,
      title={UNet++: A Nested U-Net Architecture for Medical Image Segmentation}, 
      author={Zongwei Zhou and Md Mahfuzur Rahman Siddiquee and Nima Tajbakhsh and Jianming Liang},
      year={2018},
      eprint={1807.10165},
      archivePrefix={arXiv},
      primaryClass={cs.CV},
      url={https://arxiv.org/abs/1807.10165}, 
}

@misc{unet,
      title={U-Net: Convolutional Networks for Biomedical Image Segmentation}, 
      author={Olaf Ronneberger and Philipp Fischer and Thomas Brox},
      year={2015},
      eprint={1505.04597},
      archivePrefix={arXiv},
      primaryClass={cs.CV},
      url={https://arxiv.org/abs/1505.04597}, 
}

@misc{transunet,
      title={TransUNet: Transformers Make Strong Encoders for Medical Image Segmentation}, 
      author={Jieneng Chen and Yongyi Lu and Qihang Yu and Xiangde Luo and Ehsan Adeli and Yan Wang and Le Lu and Alan L. Yuille and Yuyin Zhou},
      year={2021},
      eprint={2102.04306},
      archivePrefix={arXiv},
      primaryClass={cs.CV},
      url={https://arxiv.org/abs/2102.04306}, 
}

@misc{nnunet,
      title={nnU-Net: Self-adapting Framework for U-Net-Based Medical Image Segmentation}, 
      author={Fabian Isensee and Jens Petersen and Andre Klein and David Zimmerer and Paul F. Jaeger and Simon Kohl and Jakob Wasserthal and Gregor Koehler and Tobias Norajitra and Sebastian Wirkert and Klaus H. Maier-Hein},
      year={2018},
      eprint={1809.10486},
      archivePrefix={arXiv},
      primaryClass={cs.CV},
      url={https://arxiv.org/abs/1809.10486}, 
}

@misc{LimitReview,
      title={Medical Image Segmentation with Limited Supervision: A Review of Deep Network Models}, 
      author={Jialin Peng and Ye Wang},
      year={2021},
      eprint={2103.00429},
      archivePrefix={arXiv},
      primaryClass={cs.CV},
      url={https://arxiv.org/abs/2103.00429}, 
}

@article{large-semi,
  title={Local label propagation for large-scale semi-supervised learning},
  author={Zhuang, Chengxu and Ding, Xuehao and Murli, Divyanshu and Yamins, Daniel},
  journal={arXiv preprint arXiv:1905.11581},
  year={2019}
}

@misc{expansive_medical1,
      title={Copycats: the many lives of a publicly available medical imaging dataset}, 
      author={Amelia Jiménez-Sánchez and Natalia-Rozalia Avlona and Dovile Juodelyte and Théo Sourget and Caroline Vang-Larsen and Anna Rogers and Hubert Dariusz Zajac and Veronika Cheplygina},
      year={2024},
      eprint={2402.06353},
      archivePrefix={arXiv},
      primaryClass={cs.CV},
      url={https://arxiv.org/abs/2402.06353}, 
}

@inproceedings{sam,
  title={Segment anything},
  author={Kirillov, Alexander and Mintun, Eric and Ravi, Nikhila and Mao, Hanzi and Rolland, Chloe and Gustafson, Laura and Xiao, Tete and Whitehead, Spencer and Berg, Alexander C and Lo, Wan-Yen and others},
  booktitle={Proceedings of the IEEE/CVF international conference on computer vision},
  pages={4015--4026},
  year={2023}
}

@inproceedings{crossconsistent,
  title={Cross prompting consistency with segment anything model for semi-supervised medical image segmentation},
  author={Miao, Juzheng and Chen, Cheng and Zhang, Keli and Chuai, Jie and Li, Quanzheng and Heng, Pheng-Ann},
  booktitle={International Conference on Medical Image Computing and Computer-Assisted Intervention},
  pages={167--177},
  year={2024},
  organization={Springer}
}

@article{fan2023stablesegmentmodel,
  title={Stable segment anything model},
  author={Fan, Qi and Tao, Xin and Ke, Lei and Ye, Mingqiao and Zhang, Yuan and Wan, Pengfei and Wang, Zhongyuan and Tai, Yu-Wing and Tang, Chi-Keung},
  journal={arXiv preprint arXiv:2311.15776},
  year={2023}
}

@article{cheng2023sam,
  title={Sam on medical images: A comprehensive study on three prompt modes},
  author={Cheng, Dongjie and Qin, Ziyuan and Jiang, Zekun and Zhang, Shaoting and Lao, Qicheng and Li, Kang},
  journal={arXiv preprint arXiv:2305.00035},
  year={2023}
}

@article{Crohn_D,
author = {Ding, Honglei and Jiang, Kefang and Gao, Chen and Lu, Liangji and Zhang, Huani and Chen, Haibo and Gao, Xuning and Zhou, Kefeng and Sun, Zhichao},
year = {2022},
month = {07},
pages = {},
title = {Assessing the inflammatory severity of the terminal ileum in Crohn disease using radiomics based on MRI},
volume = {22},
journal = {BMC Medical Imaging},
doi = {10.1186/s12880-022-00844-z}
}

@article{radiologist,
author = {Chupetlovska, Kalina and Akinci D'Antonoli, Tugba and Bodalal, Zuhir and Abdelatty, Mohamed and Erenstein, Hendrik and Santinha, João and Huisman, Merel and Visser, Jacob and Trebeschi, Stefano and Groot Lipman, Kevin},
year = {2025},
month = {05},
pages = {},
title = {ESR Essentials: a step-by-step guide of segmentation for radiologists-practice recommendations by the European Society of Medical Imaging Informatics},
journal = {European radiology},
doi = {10.1007/s00330-025-11621-1}
}

@book{nextframe1,
    author = {Isenberg, David A and Renton, Peter},
    title = {Imaging in Rheumatology},
    publisher = {Oxford University Press},
    year = {2002},
    month = {12},
    isbn = {9780192632630},
    doi = {10.1093/oso/9780192632630.001.0001},
    url = {https://doi.org/10.1093/oso/9780192632630.001.0001},
}

@ARTICLE{nextframe2,
       author = {{Tian}, Fengrui and {Tian}, Zhiqiang and {Chen}, Zhang and {Zhang}, Dong and {Du}, Shaoyi},
        title = "{Surface-GCN: Learning interaction experience for organ segmentation in 3D medical images}",
      journal = {Medical Physics},
         year = 2023,
        month = aug,
       volume = {50},
       number = {8},
        pages = {5030-5044},
          doi = {10.1002/mp.16280},
       adsurl = {https://ui.adsabs.harvard.edu/abs/2023MedPh..50.5030T}
}

@article{medsam,
   title={Segment anything in medical images},
   volume={15},
   ISSN={2041-1723},
   url={http://dx.doi.org/10.1038/s41467-024-44824-z},
   DOI={10.1038/s41467-024-44824-z},
   number={1},
   journal={Nature Communications},
   publisher={Springer Science and Business Media LLC},
   author={Ma, Jun and He, Yuting and Li, Feifei and Han, Lin and You, Chenyu and Wang, Bo},
   year={2024},
   month=jan }

@misc{medsamadapter,
      title={Medical SAM Adapter: Adapting Segment Anything Model for Medical Image Segmentation}, 
      author={Junde Wu and Wei Ji and Yuanpei Liu and Huazhu Fu and Min Xu and Yanwu Xu and Yueming Jin},
      year={2023},
      eprint={2304.12620},
      archivePrefix={arXiv},
      primaryClass={cs.CV},
      url={https://arxiv.org/abs/2304.12620}, 
}

@misc{masam,
      title={MA-SAM: Modality-agnostic SAM Adaptation for 3D Medical Image Segmentation}, 
      author={Cheng Chen and Juzheng Miao and Dufan Wu and Zhiling Yan and Sekeun Kim and Jiang Hu and Aoxiao Zhong and Zhengliang Liu and Lichao Sun and Xiang Li and Tianming Liu and Pheng-Ann Heng and Quanzheng Li},
      year={2023},
      eprint={2309.08842},
      archivePrefix={arXiv},
      primaryClass={cs.CV},
      url={https://arxiv.org/abs/2309.08842}, 
}

@article{lora,
  title={Lora: Low-rank adaptation of large language models.},
  author={Hu, Edward J and Shen, Yelong and Wallis, Phillip and Allen-Zhu, Zeyuan and Li, Yuanzhi and Wang, Shean and Wang, Lu and Chen, Weizhu and others},
  journal={ICLR},
  volume={1},
  number={2},
  pages={3},
  year={2022}
}

@article{FUMPE,
  author       = {Masoudi, Mojtaba and Pourreza, Hamid‐Reza and
                  Saadatmand‐Tarzjan, Mahdi and Eftekhari, Noushin and
                  Shafiee Zargar, Fateme and Pezeshki Rad, Masoud},
  title        = {A New Dataset of Computed‑Tomography Angiography Images
                  for Computer‑Aided Detection of Pulmonary Embolism},
  journal      = {Scientific Data},
  volume       = {5},
  number       = {180180},
  pages        = {1--9},
  year         = {2018},
  month        = sep,
  doi          = {10.1038/sdata.2018.180},
  pmid         = {30179235},
  pmcid        = {PMC6122162}
}

@inproceedings{howtoprompt,
  title={Benchmarking human and automated prompting in the segment anything model},
  author={Quesada, Jorge and Fowler, Zoe and Alotaibi, Mohammad and Prabhushankar, Mohit and AlRegib, Ghassan},
  booktitle={2024 IEEE International Conference on Big Data (BigData)},
  pages={1625--1634},
  year={2024},
  organization={IEEE}
}

@article{howtoprompt2,
  title={Learning to prompt segment anything models},
  author={Huang, Jiaxing and Jiang, Kai and Zhang, Jingyi and Qiu, Han and Lu, Lewei and Lu, Shijian and Xing, Eric},
  journal={arXiv preprint arXiv:2401.04651},
  year={2024}
}

@article{huang2024segment,
  title={Segment anything model for medical images?},
  author={Huang, Yuhao and Yang, Xin and Liu, Lian and Zhou, Han and Chang, Ao and Zhou, Xinrui and Chen, Rusi and Yu, Junxuan and Chen, Jiongquan and Chen, Chaoyu and others},
  journal={Medical Image Analysis},
  volume={92},
  pages={103061},
  year={2024},
  publisher={Elsevier}
}

@inproceedings{models2,
  title={Models genesis: Generic autodidactic models for 3d medical image analysis},
  author={Zhou, Zongwei and Sodha, Vatsal and Rahman Siddiquee, Md Mahfuzur and Feng, Ruibin and Tajbakhsh, Nima and Gotway, Michael B and Liang, Jianming},
  booktitle={International conference on medical image computing and computer-assisted intervention},
  pages={384--393},
  year={2019},
  organization={Springer}
}

@inproceedings{catnet,
  title={Few shot medical image segmentation with cross attention transformer},
  author={Lin, Yi and Chen, Yufan and Cheng, Kwang-Ting and Chen, Hao},
  booktitle={International Conference on Medical Image Computing and Computer-Assisted Intervention},
  pages={233--243},
  year={2023},
  organization={Springer}
}

@misc{autosam,
      title={AutoSAM: Adapting SAM to Medical Images by Overloading the Prompt Encoder}, 
      author={Tal Shaharabany and Aviad Dahan and Raja Giryes and Lior Wolf},
      year={2023},
      eprint={2306.06370},
      archivePrefix={arXiv},
      primaryClass={cs.CV},
      url={https://arxiv.org/abs/2306.06370}, 
}

@misc{sammed2d,
      title={SAM-Med2D}, 
      author={Junlong Cheng and Jin Ye and Zhongying Deng and Jianpin Chen and Tianbin Li and Haoyu Wang and Yanzhou Su and Ziyan Huang and Jilong Chen and Lei Jiang and Hui Sun and Junjun He and Shaoting Zhang and Min Zhu and Yu Qiao},
      year={2023},
      eprint={2308.16184},
      archivePrefix={arXiv},
      primaryClass={cs.CV},
      url={https://arxiv.org/abs/2308.16184}, 
}

@inproceedings{zhang2020characterizing,
  title={Characterizing label errors: confident learning for noisy-labeled image segmentation},
  author={Zhang, Minqing and Gao, Jiantao and Lyu, Zhen and Zhao, Weibing and Wang, Qin and Ding, Weizhen and Wang, Sheng and Li, Zhen and Cui, Shuguang},
  booktitle={International conference on medical image computing and computer-assisted intervention},
  pages={721--730},
  year={2020},
  organization={Springer}
}

@article{song2022learning,
  title={Learning from noisy labels with deep neural networks: A survey},
  author={Song, Hwanjun and Kim, Minseok and Park, Dongmin and Shin, Yooju and Lee, Jae-Gil},
  journal={IEEE transactions on neural networks and learning systems},
  volume={34},
  number={11},
  pages={8135--8153},
  year={2022},
  publisher={IEEE}
}

@inproceedings{li2024self,
  title={Self-supervised alignment learning for medical image segmentation},
  author={Li, Haofeng and Ouyang, Yiming and Wan, Xiang},
  booktitle={2024 IEEE International Symposium on Biomedical Imaging (ISBI)},
  pages={1--5},
  year={2024},
  organization={IEEE}
}

@article{xu2025mambavesselnet++,
  title={MambaVesselNet++: A Hybrid CNN-Mamba Architecture for Medical Image Segmentation},
  author={Xu, Qing and Chen, Yanming and Li, Yue and Liu, Ziyu and Lou, Zhenye and Zhang, Yixuan and He, Xiangjian},
  journal={arXiv preprint arXiv:2507.19931},
  year={2025}
}

@article{yang2023keypoint,
  title={Keypoint-augmented self-supervised learning for medical image segmentation with limited annotation},
  author={Yang, Zhangsihao and Ren, Mengwei and Ding, Kaize and Gerig, Guido and Wang, Yalin},
  journal={Advances in Neural Information Processing Systems},
  volume={36},
  pages={60724--60747},
  year={2023}
}

@article{tang2025few,
  title={Few-shot medical image segmentation with high-fidelity prototypes},
  author={Tang, Song and Yan, Shaxu and Qi, Xiaozhi and Gao, Jianxin and Ye, Mao and Zhang, Jianwei and Zhu, Xiatian},
  journal={Medical Image Analysis},
  volume={100},
  pages={103412},
  year={2025},
  publisher={Elsevier}
}

@inproceedings{ding2023few,
  title={Few-shot medical image segmentation with cycle-resemblance attention},
  author={Ding, Hao and Sun, Changchang and Tang, Hao and Cai, Dawen and Yan, Yan},
  booktitle={Proceedings of the IEEE/CVF winter conference on applications of computer vision},
  pages={2488--2497},
  year={2023}
}

@article{karimi2020deep,
  title={Deep learning with noisy labels: Exploring techniques and remedies in medical image analysis},
  author={Karimi, Davood and Dou, Haoran and Warfield, Simon K and Gholipour, Ali},
  journal={Medical image analysis},
  volume={65},
  pages={101759},
  year={2020},
  publisher={Elsevier}
}

@article{unetmanba,
  title={Vm-unet: Vision mamba unet for medical image segmentation},
  author={Ruan, Jiacheng and Li, Jincheng and Xiang, Suncheng},
  journal={arXiv preprint arXiv:2402.02491},
  year={2024}
}

@inproceedings{scarcefonda,
  title={Navigating data scarcity using foundation models: A benchmark of few-shot and zero-shot learning approaches in medical imaging},
  author={Woerner, Stefano and Baumgartner, Christian F},
  booktitle={International Workshop on Foundation Models for General Medical AI},
  pages={30--39},
  year={2024},
  organization={Springer}
}

@inproceedings{scribble,
  title={Scribblevc: Scribble-supervised medical image segmentation with vision-class embedding},
  author={Li, Zihan and Zheng, Yuan and Luo, Xiangde and Shan, Dandan and Hong, Qingqi},
  booktitle={Proceedings of the 31st ACM International Conference on Multimedia},
  pages={3384--3393},
  year={2023}
}

@inproceedings{wu2023self,
  title={Self-prompting large vision models for few-shot medical image segmentation},
  author={Wu, Qi and Zhang, Yuyao and Elbatel, Marawan},
  booktitle={MICCAI workshop on domain adaptation and representation transfer},
  pages={156--167},
  year={2023},
  organization={Springer}
}

@inproceedings{radford2021learning,
  title={Learning transferable visual models from natural language supervision},
  author={Radford, Alec and Kim, Jong Wook and Hallacy, Chris and Ramesh, Aditya and Goh, Gabriel and Agarwal, Sandhini and Sastry, Girish and Askell, Amanda and Mishkin, Pamela and Clark, Jack and others},
  booktitle={International conference on machine learning},
  pages={8748--8763},
  year={2021},
  organization={PmLR}
}

@inproceedings{liu2022adaptive,
  title={Adaptive early-learning correction for segmentation from noisy annotations},
  author={Liu, Sheng and Liu, Kangning and Zhu, Weicheng and Shen, Yiqiu and Fernandez-Granda, Carlos},
  booktitle={Proceedings of the IEEE/CVF conference on computer vision and pattern recognition},
  pages={2606--2616},
  year={2022}
}

@misc{kits,
      title={The KiTS21 Challenge: Automatic segmentation of kidneys, renal tumors, and renal cysts in corticomedullary-phase CT}, 
      author={Nicholas Heller and Fabian Isensee and Dasha Trofimova and Resha Tejpaul and Zhongchen Zhao and Huai Chen and Lisheng Wang and Alex Golts and Daniel Khapun and Daniel Shats and Yoel Shoshan and Flora Gilboa-Solomon and Yasmeen George and Xi Yang and Jianpeng Zhang and Jing Zhang and Yong Xia and Mengran Wu and Zhiyang Liu and Ed Walczak and Sean McSweeney and Ranveer Vasdev and Chris Hornung and Rafat Solaiman and Jamee Schoephoerster and Bailey Abernathy and David Wu and Safa Abdulkadir and Ben Byun and Justice Spriggs and Griffin Struyk and Alexandra Austin and Ben Simpson and Michael Hagstrom and Sierra Virnig and John French and Nitin Venkatesh and Sarah Chan and Keenan Moore and Anna Jacobsen and Susan Austin and Mark Austin and Subodh Regmi and Nikolaos Papanikolopoulos and Christopher Weight},
      year={2023},
      eprint={2307.01984},
      archivePrefix={arXiv},
      primaryClass={cs.CV}
}

@misc{refiner,
      title={SAMRefiner: Taming Segment Anything Model for Universal Mask Refinement}, 
      author={Yuqi Lin and Hengjia Li and Wenqi Shao and Zheng Yang and Jun Zhao and Xiaofei He and Ping Luo and Kaipeng Zhang},
      year={2025},
      eprint={2502.06756},
      archivePrefix={arXiv},
      primaryClass={cs.CV},
      url={https://arxiv.org/abs/2502.06756}, 
}

@article{sam2,
  title={Sam 2: Segment anything in images and videos},
  author={Ravi, Nikhila and Gabeur, Valentin and Hu, Yuan-Ting and Hu, Ronghang and Ryali, Chaitanya and Ma, Tengyu and Khedr, Haitham and R{\"a}dle, Roman and Rolland, Chloe and Gustafson, Laura and others},
  journal={arXiv preprint arXiv:2408.00714},
  year={2024}
}

@article{medsam2,
  title={Medsam2: Segment anything in 3d medical images and videos},
  author={Ma, Jun and Yang, Zongxin and Kim, Sumin and Chen, Bihui and Baharoon, Mohammed and Fallahpour, Adibvafa and Asakereh, Reza and Lyu, Hongwei and Wang, Bo},
  journal={arXiv preprint arXiv:2504.03600},
  year={2025}
}
